%% file: l2s_nips16-arxiv.tex
\documentclass[oneside,english]{article}

\usepackage[numbers]{natbib}

\usepackage[noend]{algorithmic}
\usepackage{booktabs}
\input{preamble}

\usepackage{tikz}
\usetikzlibrary{shapes.gates.logic.US,trees,positioning,arrows,automata,decorations.pathmorphing,chains}
\usepackage[final]{nips_2016}
\newcommand{\newalgorithmnofloat}[3]{%
  \begin{algorithm}[H]
    \caption{#2}
    \begin{small}
      \begin{algorithmic}[1]
        #3
      \end{algorithmic}
    \end{small}
    \label{alg:#1}
  \end{algorithm}}

  \author{
Kai-Wei Chang\\
University of Illinois\\
\href{mailto:Kai-Wei Chang <kw@kwchang.net>} {\nolinkurl{kw@kwchang.net}}
\And
He He\\
University of Maryland\\
\href{mailto:He He <hhe@cs.umd.edu>}{\nolinkurl{hhe@cs.umd.edu}}
\And
Hal Daum\'e III\\
University of Maryland\\
\href{mailto:Hal Daume III <me@hal3.name>}{\nolinkurl{me@hal3.name}}
\AND
John Langford\\
Microsoft Research\\
\href{mailto:John Langford <jcl@microsoft.com>}{\nolinkurl{jcl@microsoft.com}}
\And
Stephane Ross\\
Google\\
\href{mailto:Stephane Ross <stephaneross@google.com>}{\nolinkurl{stephaneross@google.com}}
}

\usepackage{times}
\usepackage{graphicx} 
\usepackage{subfigure}

\usepackage{paralist,amsfonts} 

\usepackage{algorithmic}
\newcommand{\nn}{\textsc{Snn}}
\newcommand{\dn}{\textsc{Dyna}}
\newcommand{\red}{\textsc{RedS}}
\newcommand{\our}{\textsc{L2S}}

\hypersetup{pdftitle={A Credit Assigment Compiler for Joint Prediction},
            pdfauthor={Anonymous Submission}}

\newcommand{\myrun}{\FUN{MyRun}(...)}

\newcommand{\ignore}[1]{}

\begin{document}

\title{A Credit Assignment Compiler for Joint Prediction}
\maketitle

\begin{abstract}
	Many machine learning applications involve jointly predicting
        multiple mutually dependent output variables.  Learning to
        search is a family of methods where the complex decision
        problem is cast into a sequence of decisions via a search
        space.  Although these methods have shown promise both in
        theory and in practice, implementing them has been
        burdensomely awkward.   In this paper,
	we show the search space can be defined by an arbitrary
        imperative program, turning learning to search into a credit
        assignment compiler.  Altogether with the algorithmic
        improvements for the compiler, we radically reduce the
        complexity of programming and the running time.
        We demonstrate the feasibility of our approach on multiple joint
        prediction tasks.  In all cases, we obtain accuracies as high
        as alternative approaches, at drastically reduced execution
        and programming time.
\end{abstract} 

\section{Introduction}
\label{sec:intro}


Many applications require a predictor to make coherent 
decisions.   
As an example, consider recognizing a handwritten word where each
character might be recognized in turn to understand the word. Here, it
is commonly observed that exposing information from related
predictions (i.e. adjacent letters) aids individual
predictions. Furthermore, optimizing a joint loss function can improve
the gracefulness of error recovery.  Despite these advantages, it is
empirically common to build independent predictors, in settings where
joint prediction naturally applies, because they are simpler to
implement and faster to run.  Can we make joint prediction algorithms
as easy and fast to program and compute while maintaining their
theoretical benefits?


Methods making a sequence of sub-decisions have been proposed for
handling complex joint predictions in a variety of applications,
including sequence tagging~\cite{mccallum00memm}, dependency parsing (known as 
transition-based method)~\cite{nivre03parsing}, 
machine translation~\cite{germann03decoding}, and co-reference resolution~\cite{soon01machine}.
Recently, general search-based joint prediction approaches 
(e.g., \cite{collins04incremental,daume05laso,doppa14hcsearch,huang12structured,ross11dagger})
have been investigated.  
The key issue of these search-based approaches is credit assignment: When something goes wrong
do you blame the first, second, or third prediction? 
Existing methods often take two strategies:
\begin{itemize}
	\item The system ignores the possibility that a previous
          prediction may have been wrong, different costs have
          different errors, or the difference between train-time and
          test-time prediction.
\item  The system may use handcrafted credit assignment
heuristics to cope with errors that the underlying algorithm makes and the long-term
outcomes of decisions.  
\end{itemize}
Both approaches may lead to statistical inconsistency: when features are not
rich enough for perfect prediction, the machine learning may converge sub-optimally. 

In contrast, learning to search
approaches~\cite{chang15lols,daume09searn,ross14aggrevate}
automatically handle the credit assignment problem by decomposing the
production of the joint output in terms of an explicit search space
(states, actions, etc.); and learning a control policy that takes
actions in this search space.  These have formal correctness
guarantees which differ qualitatively from graphical models such as
the Conditional Random Fields~\cite{lafferty01crf} and structured SVMs
\cite{taskar03mmmn,tsochantaridis04svmiso}.  Despite the good properties, none of these methods
have been widely adopted because the specification of a search space
as a finite state machine is awkward and naive implementations do not
fully demonstrate the ability of these methods.

\newalgorithm{sequential}%
{\FUN{MyRun}(\VAR{X}) \% for sequence tagging, X: input sequence, Y: output} 
  {
    \SETST{{Y}}{[]}
    \FOR{{\VAR{t}} = \CON{1} to \FUN{len}(\VAR{X})}
      \SETST{ref}{\VAR{X}{[}\VAR{t}{]}\texttt{.}{true\_label}}
      \STATE{\VAR{Y}[\VAR{t}] $\leftarrow$ \FUN{predict}(\texttt{\footnotesize x=}\VAR{examples}{[}{t}{]}, \texttt{\footnotesize y=}\VAR{ref}, \texttt{\footnotesize tag=}\VAR{t}, \texttt{\footnotesize condition=}[1:\VAR{t}-1])}
    \ENDFOR
    \STATE \FUN{loss}(number of \VAR{Y}[\VAR{t}] $\neq$ \VAR{X}[\VAR{t}]\texttt{.}true\_label)
    \STATE \textbf{return} \VAR{Y}
  }


In this paper, we cast learning to search into a credit assignment
compiler with a new programming abstraction for representing a search
space. Together with several algorithmic improvements, this radically
reduces both the complexity of programming and the running time.  The
programming interface has the following advantages:
\begin{compactitem}
\item The same decoding function (see Alg.  \ref{alg:sequential} and
  Sec.  \ref{sec:interface} for example) is used for training and
  prediction so a developer need only code desired test time behavior
  and gets training ``for free. '' This simple implementation prevents
  common train/test asynchrony bugs.
\item The compiler automatically ensures the model learns to avoid
  compounding errors and makes a sequence of coherent decisions.
\item The library functions are in a reduction stack so as base classifiers
  and learning to search approaches improve, so does joint prediction
  performance.
\end{compactitem}

Without extra implementation cost\footnote{In fact, with library
  supports, developing a new task often requires only a few lines of
  code.}, implementations enabled by the credit assignment compiler
achieve outstanding empirical performance both in accuracy and in
speed.  This provides strong simple baselines for future research and
demonstrates the compiler approach to solving complex prediction
problems may be of broad interest.\footnote{Experiments and
  implementations will be released.}

\section{Programmable Learning to Search}
\label{sec:interface}

We first describe the proposed programmable joint prediction paradigm.
Algorithm \ref{alg:sequential} shows sample code for a part of speech
tagger (or generic sequence labeler) under Hamming loss.  The
algorithm takes as input a sequence of examples (e.g., words), and
predicts the meaning of each element in turn. The $i$th prediction
depends on previous predictions.\footnote{In this example, we use
 the library's support for generating implicit features based on
  previous predictions.}  It uses two underlying library functions,
\predict\ and \loss.  The function \predict\ returns individual
predictions based on $x$ while \loss\ allows the declaration of an
\emph{arbitrary} loss for the point set of predictions.
The \loss\ function and the reference $y$ inputted to \predict\
are only used in the training phase and it has no effect in the test phase.
Surprisingly, this single library interface is sufficient for both
testing \emph{and training}, when augmented to include label ``advice'' from a training
set as a reference decision (by the parameter $y$).
This means that a developer only has to specify the desired \emph{test time behavior} and gets training with minor additional decoration.
The underlying system works as a credit assignment compiler to translate the user-specified decoding function and labeled data
into updates of the learning model.  

How can you learn a good \FUN{predict} function given just an
imperative program like Algorithm \ref{alg:sequential}?  In the
following, we show that it is essential to run the \myrun\ function 
(e.g., Algorithm \ref{alg:sequential})
many times, ``trying out'' different
versions of \predict\ to learn one that yields low \loss.  We
begin with formal definitions of joint prediction and a search space.

\begin{figure}[t]
\centering
\noindent\begin{tabular}{|@{}p{0.3\linewidth}|p{0.65\linewidth}|}
\hline
\includegraphics[width=\linewidth]{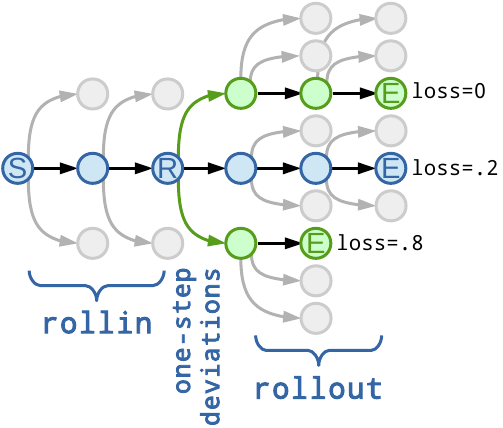} & 
\vspace{-1.2in}
The system begins at the start state $S$ and chooses the middle action twice according to the \textbsf{rollin} policy. At state $R$ it considers both 
the chosen action (middle) and one-step deviations from that action (top 
and bottom). Each of these deviations is completed using the \textbsf{rollout} 
policy until reaching an end state, at which point the loss is collected. 
Here, we learn that deviating to the top action (instead of middle) at state 
$R$ decreases the loss by $0.2$.\\
\hline
\end{tabular}
\caption{A search space implicitly defined by an imperative program.}
\label{fig:searchspace}
\end{figure}

\paragraph{Joint Prediction.  }
Joint prediction aims to induce a function $f$
such that for any $X \in \mathcal{X}$ (the input space), $f$ produces an
output $f(X) = Y \in \mathcal{Y}(X)$ in a (possibly input-dependent) space
$\mathcal{Y}(X)$. The output $Y$ often can be decomposed into
smaller pieces (e.g., $y_1, y_2,\ldots$), which are tied together by features,
by a loss function and/or by statistical dependence.
There is a task-specific loss function $\ell : \mathcal{Y} \times \mathcal{Y}
\rightarrow \mathbb{R}^{\geq0}$, where $\ell(Y^*,\hat Y)$ tells us how bad it is to predict
$\hat Y$ when the true is $Y^*$.  

\paragraph{Search Space.} In our framework, the joint variable $\hat Y$
is produced incrementally by traversing a search space, which 
is defined by states $s\in S$ and a mapping
$A:S\rightarrow2^{S}$ defining the set of valid next 
states.\footnote{Comprehensive strategies for defining search space have been 
discussed 
\citep{doppa14hcsearch}. The theoretical properties do not depend on which search space definition is used.}
One of the states is a unique start state $S$ while some of the others are
end states $e\in E$. Each end state corresponds to some output variable $Y_e$.  
The goal of learning is finding a function $f:X_{s}\rightarrow S$ that
uses the features of an input state ($x_s$) to choose the next state
so as to minimize the loss $\ell(Y^*,Y_e)$ on a holdout test set.\footnote{Note that we 
use $X$ and $Y$ to represent joint input and output and use $x$ and $y$ to 
represent input and output to function $f$ and \FUN{PREDICT}.}  Follow
reinforcement learning terminology, we call the function a policy and
call the learned function $f$ a learned policy $\pi_f$.

\paragraph{Turning Search Space into an Imperative Program}

\begin{figure}[t]
\centering
\noindent\begin{tabular}{|@{}p{0.58\linewidth}|p{0.37\linewidth}|}
\hline
\begin{minipage}[t]{\linewidth}
	{
	\vspace{3pt}
\setdefaultleftmargin{2em}{2em}{2em}{2em}{2em}{2em}
\ The definition of a TDOLR program: 
	\begin{compactitem}
	\item  Always terminate.
	\item  Takes as input any relevant feature information $X$.
	\item  Make zero or more calls to an oracle $O:X'\rightarrow Y$ which provides
a discrete outcome. 
	\item  Report a loss $L$ on termination.
	\end{compactitem}
	}
\end{minipage}
& 
\vspace{-10pt}
\begin{minipage}[t]{\linewidth}
\newalgorithmnofloat{TDOLR}%
  {\FUN{TDOLR}(\VAR{X})}%
  {
    \SETST{\VAR{s}}{\VAR{a}}
    \WHILE{\VAR{s} $\not\in$ \VAR{E}}
      \STATE Compute \VAR{$x_s$} from \VAR{X} and \VAR{s}
      \SETST{s}{\FUN{O}(\VAR{$x_s$})}
    \ENDWHILE
    \RETURN{\FUN{Loss}(\VAR{s})}
  }  
 \vspace{3pt}
\end{minipage}\\
\hline
\end{tabular}
\caption{Left: the definition; right: A TDOLR program simulates the search 
space.  } 
\label{fig:tdolr}
\end{figure}

Surprisingly, search space can be represented by a class of 
imperative program, called Terminal Discrete Oracle Loss
Reporting (TDOLR) programs.   The formal definition of TDOLR is listed in Figure \ref{fig:tdolr}.
Without loss of generality, we assume the number of choices is fixed in a 
search space, and the following theorem holds:
\begin{theorem}
For every TDOLR program, there exist an equivalent search space and
for every search space there exists an equivalent TDOLR program.\end{theorem}

\begin{proof}
A search space is defined by $(A,E,S,l)$. We show there is a TDOLR
program which can simulate the search space in algorithm \ref{alg:TDOLR}.
This algorithm does a straightforward execution of the search space,
followed by reporting of the loss on termination. This completes the
second claim. 

For the first claim, we need to define, $(A,E,S,l)$ given a TDOLR
program such that the search space can simulate the TDOLR program.
At any point in the execution of TDOLR, we define an equivalent state
$s=(O(X_{1}),...,O(X_{n}))$ where $n$ is the number of calls to
the oracle. We define $a$ as the sequence of zero length, and we
define $E$ as the set of states after which TDOLR terminates. For
each $s\in E$ we define $l(s)$ as the loss reported on termination.
This search space manifestly outputs the same loss as the TDOLR program. 
\end{proof}

The practical implication of this theorem is that instead of
specifying search spaces, we can specify a TDOLR program (such as
Algorithm \ref{alg:sequential}), radically reducing the programming
complexity of joint prediction.



\section{Credit Assignment Compiler for Training Joint Predictor} \label{sec:comp}
Now, we show how credit assignment compiler turn a TDOLR program and training 
data into model updates.   
In the training phase, the supervised signals are used in two places:
1) to define the loss function, and 2) to construct a reference policy $\pi^{*}$.
The reference policy returns at any prediction point a ``suggestion'' as to a good next state.\footnote{Some papers assume the reference policy is optimal. An optimal policy always chooses the best next state assuming it gets to make all future decisions as well.}
The general strategy is, for some number of epochs, and for each example $(X,Y)$ in the training data, to do the following:
{
\setdefaultleftmargin{2em}{2em}{2em}{2em}{2em}{2em}
\begin{compactenum}
\item Execute \myrun\ on $X$ with a \textbsf{rollin policy} to obtain a trajectory of actions $\vec a$ and loss $\ell_0$
\item Many times:
\begin{compactenum}
\item For some (or for all) time step $t \leq |\vec a|$
\item For some (or for all) alternative action $a_t' \neq a_t$  ($a_t$ is the action taken 
	by $\vec a$ in time step $t$)
\item Execute \myrun\  on $X$, with \FUN{predict} returning $a_{1:t-1}$ initially, then $a_t'$, then acting according to a \textbsf{rollout policy} to obtain a new loss $\ell_{t,a_t'}$
\item Compare the overall losses $\ell_{t,a_t}$ and $\ell_{t,a_t'}$ to 
	construct a classification/regression example that demonstrates how much 
	better or worse $a_t'$ is than $a_t$ in this context.  
\end{compactenum}
\item Update the learned policy
\end{compactenum}
}

The rollin and rollout policies can be the reference $\pi^*$, the current 
classifier $\pi_f$ or a mixture between them.  
By varying them and the manner in which classification/regression examples are 
created, this general framework can mimic algorithms like \textsc{Searn}~\cite{daume09searn}, 
 \textsc{DAgger}~\cite{ross11dagger},  \textsc{AggreVaTe}~\cite{ross14aggrevate}, and \textsc{LOLS}~\cite{chang15lols}.\footnote{E.g., rollin in LOLS is $\pi_f$ 
and rollout is a stochastic interpolation of $\pi_f$ and oracle $\pi^*$ constructed by \VAR{y}.} 

\newalgorithm{learning}%
  {\FUN{Learn}(\VAR{X}, \FUN{f})}
  {
	  \SETST{T, ex, cache}{\CON{0}, [], []}	 
	\label{learn:initbegin}
    \STATE define \FUN{Predict}(\VAR{x}, \VAR{y}) := \{ \VAR{T}\texttt{++} ; 
	\VAR{ex}[\VAR{T}-1] $\leftarrow$ \VAR{x}; \VAR{cache}[\VAR{T}-1] $\leftarrow$ 
	\FUN{f}(\VAR{x}, \VAR{y}, \textsf{\footnotesize rollin}) ;  \myreturn 
	\VAR{cache}[\VAR{T}-1]  \ \} 
	\STATE define \FUN{Loss}(\VAR{l}) := \textsf{\footnotesize no-op} 
    \STATE \FUN{MyRun}(X) \label{learn:initend}
    \FOR{\VAR{$t_0$} $=\CON{1}$ \TO \VAR{T}} \label{learn:loopt}
	\SETST{losses, t}{$\langle$\CON{0}, \CON{0}, \dots, \CON{0}$\rangle$, \CON{0}}
      \FOR{\VAR{$a_0$} $=\CON{1}$ \TO \FUN{A}(\VAR{ex}[\VAR{$t_0$}])} \label{learn:loopa}
        \STATE Define \FUN{Predict}(\VAR{x}, \VAR{y}) := \{
            \VAR{t}\texttt{++} ; 
            \myreturn
            $\brack{ 
              \textnormal{\VAR{cache}$[$\VAR{t}-1$]$} & \textnormal{if } t < t_0 \\
              \textnormal{\VAR{$a_0$}} & \textnormal{if } t = t_0 \\ 
              \textnormal{\FUN{f}(\VAR{x},\VAR{y},\textsf{\footnotesize rollout})} & \textnormal{if } t > t_0
            }$ \}
        \STATE Define \FUN{Loss}(\VAR{val}) := \{ \VAR{losses}[\VAR{$a_0$}] \texttt{+=} \VAR{val} \} \label{learn:loss}
        \STATE \FUN{MyRun}(X) \label{learn:rerun}
      \ENDFOR
      \STATE Online update with cost-sensitive example (\VAR{ex}[\VAR{$t_0$}], \VAR{losses}) \label{learn:example}
    \ENDFOR
  }

The full learning algorithm (for a single joint input \VAR{X}) is depicted in Algorithm~\ref{alg:learning}.\footnote{This algorithm is awkward because standard computational systems have a single stack. We have elected to give \FUN{MyRun} control of the stack to ease the implementation of joint prediction tasks.  Consequently, the learning algorithm does not have access to the machine stack and must be implemented as a state machine.}
In lines~\ref{learn:initbegin}--\ref{learn:initend}, a \emph{rollin} pass of 
\FUN{MyRun} is executed.  \FUN{MyRun} can generally be any TDOLR program as 
discussed (e.g., Alg. \ref{alg:sequential}). In this pass, predictions are made according to the current 
policy, \FUN{f}, flagged as rollin (this is to enable support of arbitrary 
rollin and rollout policies). Furthermore, the examples (feature vectors) 
encountered during prediction are stored in \VAR{ex}, indexed by their 
position in the sequence (\VAR{T}), and the rollin predictions are cached in 
the variable \VAR{cache} (see Sec.  \ref{sec:opt}).

The algorithm then initiates one-step deviations from this rollin trajectory. For every time step, (line~\ref{learn:loopt}), we generate a \emph{single} cost-sensitive classification example; its features are \VAR{ex}[\VAR{$t_0$}], and there are \FUN{m}(\VAR{ex}[\VAR{$t_0$}]) possible labels (=actions). For each action (line~\ref{learn:loopa}), we compute the \emph{cost} of that action by executing \FUN{MyRun} again (line~\ref{learn:rerun}) with a ``tweaked'' \FUN{Predict} which returns the cached predictions at steps before \VAR{$t_0$}, returns the perturbed action \VAR{$a_0$} at \VAR{$t_0$}, and at future timesteps calls \FUN{f} for rollouts.
The \FUN{Loss} function  accumulates the loss for the query action. Finally, a cost-sensitive classification example is generated (line~\ref{learn:example}) and fed into an online learning algorithm.  


\section{Optimizing the Credit Assignment Compiler} \label{sec:opt}
We present two algorithmic improvements which
make training orders of magnitude faster.

\paragraph{Optimization 1: Memoization}

The primary computational cost of Alg. \ref{alg:learning} is making
predictions: namely, calling the underlying classifier in Step
\ref{learn:rerun}. In order to avoid redundant predictions, we cache
previous predictions. The challenge is understanding how to know when
two predictions are going to be identical, faster than actually
computing the prediction.  To accomplish this, the user may decorate
calls to the \FUN{Predict} function with \emph{tags}.  For a graphical
models, a tag is effectively the ``name'' of a particular variable in
the graphical model. For a sequence labeling problem, the tag for a
given position might just be its index.  When calling \FUN{Predict},
the user specifies both the tag of the current prediction, and the tag
of \emph{all previous predictions} on which the current prediction
depends.  The user is guaranteeing that \emph{if} the predictions for
all the tags in the dependent variables are the same, \emph{then} the
prediction for the current example are the same.

Under this assumption, we store a cache that maps triples of
$\langle$tag, condition tags, condition predictions$\rangle$ to
$\langle$current prediction$\rangle$. The added overhead of
maintaining this data structure is tiny in comparison to making
repeated predictions on the same features.  In
line~\ref{learn:example} the learned policy changes making correctness
subtle. For data mixing algorithms (like DAgger), this potentially
changes \FUN{f$_i$} implying the memoized predictions may no longer be
up-to-date. Thus this optimization is okay \emph{if} the policy does
not change much. We evaluate this empirically in
Section~\ref{sec:pathcollapse}.

\paragraph{Optimization 2: Forced Path Collapse}

The second optimization we can use is a heuristic that only makes
rollout predictions for a constant number of steps (e.g., 2 or 4). The
intuition is that optimizing against a truly long term reward may be
impossible if features are not available at the current time
\VAR{$t_0$} which enable the underlying learner to distinguish between
the outcome of decisions far in the future. The optimization
stops rollouts after some fixed number of rollout steps.

This intuitive reasoning is correct, except for accumulating \loss. If \loss\ is only declared at the end of \FUN{MyRun}, then we must execute $T-t_0$ time steps making (possibly memoized) predictions.
However, for many problems, it is possible to declare loss \emph{early} as with Hamming loss (= number of incorrect predictions). There is no need to wait until the end of the sequence to declare a per-sequence loss: one can declare it after every prediction, and have the total loss accumulate (hence the ``\texttt{+=}'' on line~\ref{learn:loss}). We generalize this notion slightly to that of a history-independent loss:

\begin{definition}[History-independent loss]
A loss function is \emph{history-independent at state $s_0$} if, for \emph{any} final state $e$ reachable from $s_0$, and for any sequence $s_0 s_1 s_2 \dots s_i = e$: it holds that $\FUN{Loss}(e) = A(s_0) + B(s_1 s_2 \dots s_i)$, where $B$ does not depend on any state before $s_1$.
\end{definition}

For example, Hamming loss is history-independent: $A(s_0)$ corresponds
to loss through $s_0$ and $B(s_1 \dots s_i)$ is the loss after
$s_0$.\footnote{Any loss function that decomposes over the structure,
  as required by structured SVMs, is guaranteed to also be
  history-independent; the reverse is not true. Furthermore, when
  structured SVMs are run with a non-decomposable loss function, their
  runtime becomes exponential in $t$. When our approach is used with a
  loss function that's not history-independent, our runtime increases
  by a factor of $t$.}  When the loss function being optimized is
history-independent, we allow \loss\ to be declared early for this
optimization.  In addition, for tasks like transition-base dependency
parsing, although \loss\ is not decomposable over actions, expected
cost per action can be directly computed based on gold
labels~\cite{goldberg13oracles} so the array {\sl losses} can be
directly specified.

\paragraph{Speed Up} We analyize the time complexity for the sequence tagging task.
Suppose that the cost of calling the policy is $d$
and each state has $k$ actions.\footnote{Because the policy is a multiclass classifier, $d$ might hide a factor of $k$ or $\log k$.} 
Without any speed enhancements, each execution of \FUN{MyRun} takes
$\cO(T)$ time, and we execute it $Tk+1$ times, yielding an overall
complexity of $\cO(kT^2d)$ per joint example.  For comparison,
structured SVMs or CRFs with first order Markov dependencies run in
$\cO(k^2T)$ time.  When both memoization and forced path collapse are
in effect, the complexity of training drops to $\cO(Tkd)$, similar to
independent prediction.  In particular, if the $i$th prediction only
depends on the $i-1$th prediction, then at \emph{most} $Tk$ unique
predictions are made.\footnote{We use \emph{tied
    randomness}~\cite{ng00pegasus} to ensure that for any time step,
  the same policy is called.}


\section{System Performance}
\label{sec:experiments}
We present two sets of experiments. In the first set, we compare the
credit assignment compiler with existing libraries on two sequence
tagging problems: Part of Speech tagging (POS) on the Wall Street
Journal portion of the Penn Treebank; and sequence \emph{chunking}
problem: named entity recognition (NER) based on standard Begin-In-Out
encoding on the CoNLL 2003 dataset.  In the second set of experiments,
we demonstrate a simple dependency parser built by our approach
achieves strong results when comparing with system with similar
complexity. The parser is evaluated on the standard WSJ (English,
Stanford-style labels), CTB (Chinese) datasets and the CoNLL-X
datasets for 10 other languages.\footnote{PTB and CTB are prepared by
  following \cite{danqi14nndep}, and CoNLL-X is from the CoNLL shared
  task 06.}  Our approach is implemented using the Vowpal Wabbit
\cite{langford07vw} toolkit on top of a cost-sensitive classifier
\cite{beygelzimer05reductions} trained with online
updates~\cite{duchi2011adaptive,invariant,normalized}.  Details of
dataset statistics, experimental settings, additional results on other
applications, and pseudocode are in the appendix.


\subsection{Sequence Tagging Tasks}

\begin{figure}
\begin{tabular}{@{}c@{ }c@{ }c@{}}
\includegraphics[width=0.33\textwidth,clip=true,trim=24 25 38 20]{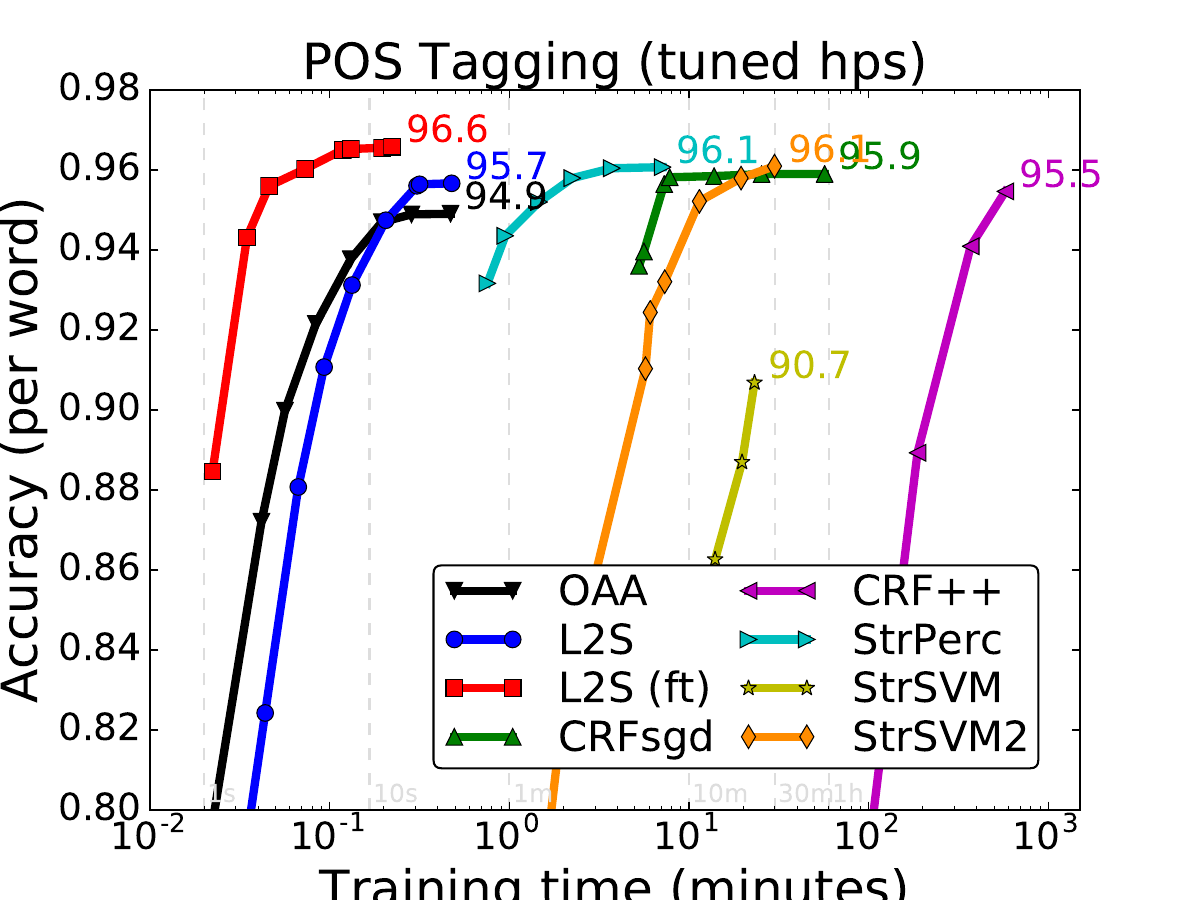}&
\includegraphics[width=0.33\textwidth,clip=true,trim=24 25 38 20]{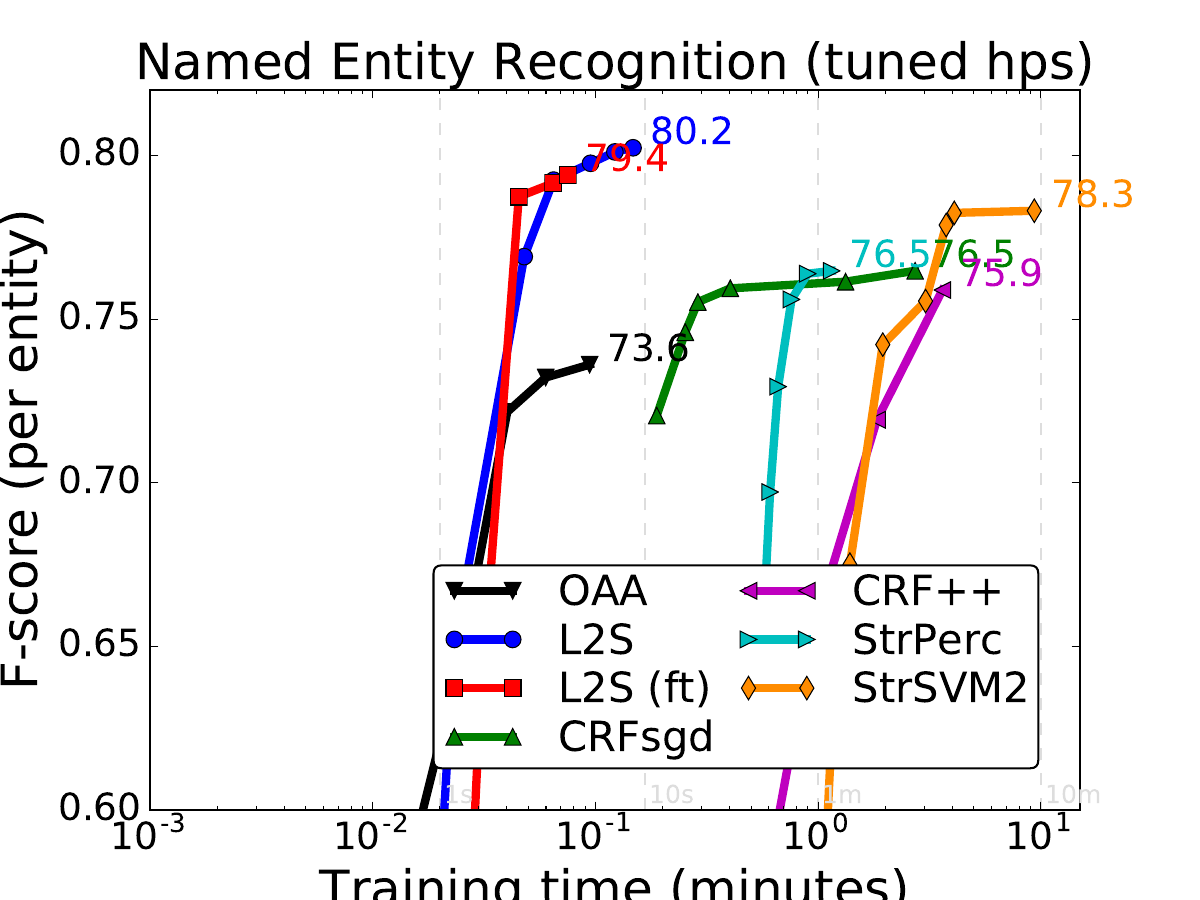}&
\includegraphics[width=0.33\linewidth,clip=true,trim=24 25 38 20]{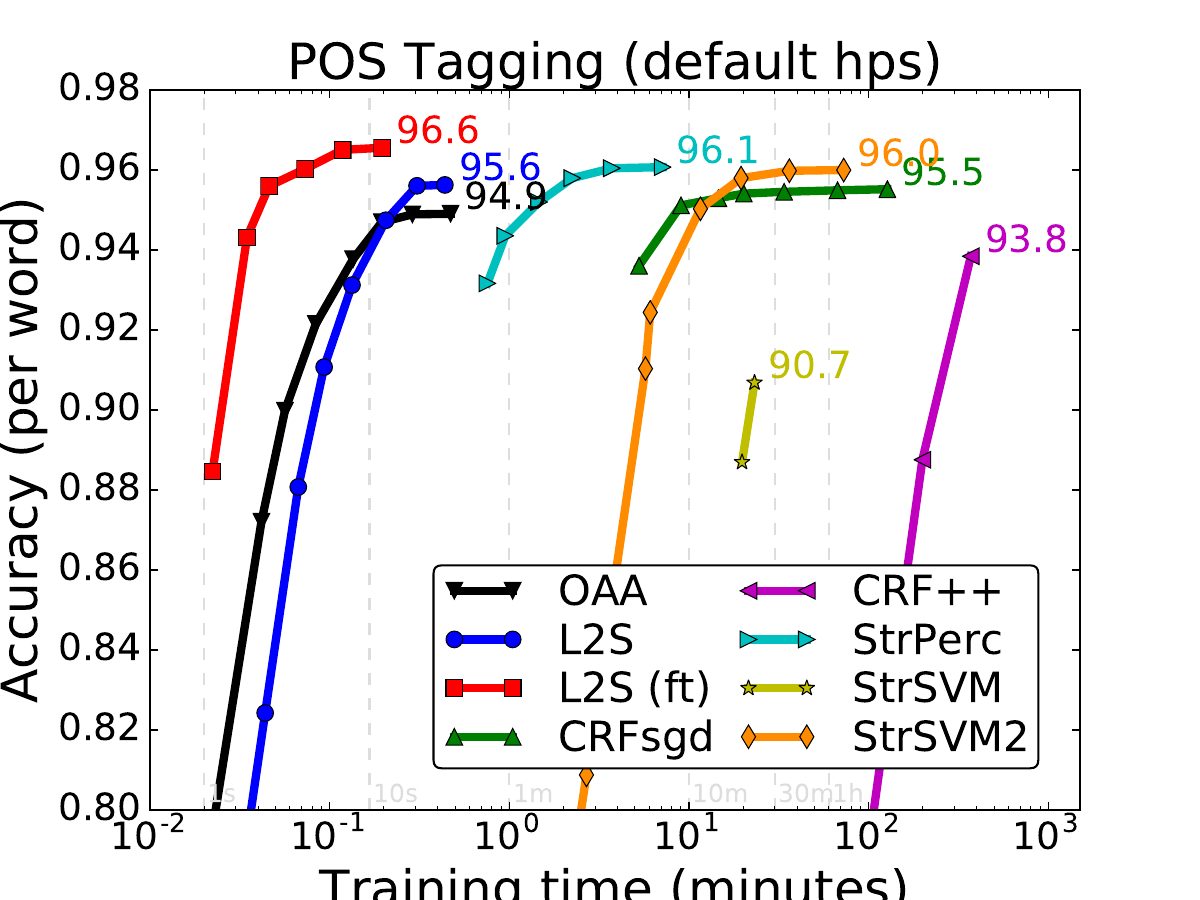}
\end{tabular}
\caption{Training time (minutes) versus test accuracy for POS and NER. 
Different points correspond to different termination 
criteria for training. The rightmost figure use \emph{default} hyperparameters
and the two left figures use hyperparameters that were tuned (for accuracy) on the holdout data.
Results of NER with default parameters are in the appendix.  
{\bf X-axis is in log scale.}
}
\label{fig:results}
\end{figure}

We compare our system with other freely available systems/algorithms, including
\system{CRF++}~\cite{crf++}, \system{CRF SGD}~\cite{crfsgd}, \system{Structured Perceptron}~\cite{collins02perceptron}, 
\system{Structured SVM}~\cite{joachims09cuttingplane}, \system{Structured SVM (DEMI-DCD)}~\cite{chang13svm}, and
an \emph{unstructured} baseline predicting each label independently, using one-against-all classification~\cite{beygelzimer05reductions}\footnote{
\system{Structured Perceptron} and \system{Structured SVM (DEMI-DCD)} are implemented in Illioins-SL\cite{CUCSR15}.
\system{DEMI-DCD} is a multi-core dual approach, while \system{Structured SVM} uses cutting-planes.
}.

For each system, we consider two situations, either the \textbf{default 
hyperparameters} or the \textbf{tuned hyperparameters} that achieved the 
best performance on holdout data. 
We report both conditions to give a sense of how sensitive each approach is to the setting of hyperparameters (the amount of hyperparameter tuning directly affects effective training time).
We use the built-in \emph{feature template} of \system{CRF++} to generate features and use them for other systems.
The templates included  neighboring words and, in the case of NER, neighboring POS tags. 
The \system{CRF++}\ templates generate $630k$ unique features for the training data.
\emph{However,} because \system{L2S}\ is also able to generate features 
from its own templates, we also provide results for \textbf{\system{L2S\ 
(ft)}} in which it uses its own feature template generation.

\paragraph{Training time.} In Figure~\ref{fig:results}, we show trade-offs between training time (x-axis, log scaled) and prediction accuracy (y-axis) for the aforementioned six systems.
For POS tagging, the independent classifier is the fastest (trains in less 
than one minute) but its performance peaks at $95\%$ accuracy. Three other 
approaches are in roughly the same time/accuracy trade-off: \system{L2s}, 
\system{L2S (ft)} and \system{Structured Perceptron}. 
\system{CRF SGD} takes about twice as long. \system{DEMI-DCD} (taking a half 
hour) and \system{CRF++} (taking over five hours) are not competitive. 
\system{Structured SVM} runs out of memory before achieving competitive 
performance (likely due to too many constraints). For NER the story is a bit 
different. The independent classifiers are not competitive. Here, the two 
variants of \system{L2S} totally dominate. In this case, \system{Structured 
Perceptron} is no longer competitive and is essentially dominated by 
\system{CRF SGD}. The only system coming close to \system{L2S}'s performance 
is \system{DEMI-DCD}, although it's performance flattens out after a few 
minutes.\footnote{We also tried giving \system{\scriptsize CRF SGD} the 
features computed by \system{L2S (ft)} on both POS and NER. On POS, its 
accuracy improved to 96.5 with essentially the same speed. On NER it's 
performance decreased.}
The trends in the runs with default hyperparameters show similar behavior to those with tuned, though some of the competing approaches suffer significantly in prediction performance. \system{Structured Perceptron} has no hyperparameters.



\paragraph{Test Time.} In addition to training time, one might care about test time behavior.
On NER, prediction times where $5.3k$ tokens/second (\system{DEMI-DCD} and \system{Structured Perceptron}, $20k$ (\system{CRF SGD} and \system{Structured SVM}), $100k$ (\system{CRF++}), $220k$ (\system{L2S (ft)}), and $285k$ (\system{L2S}). Although \system{CRF SGD} and \system{Structured Perceptron} fared well in terms of training time, their test-time behavior is suboptimal.
When the number of labels increases from $9$ (NER) to $45$ (POS) the relative advantage of \system{L2S} increases further.  The speed of \system{L2S} is about halved while for others, it is cut down by as much as a factor of $8$ due to the $O(k)$ vs $\cO(k^2)$ dependence on the label set size.  



\subsection{Dependency Parsing }
\label{sec:dep}

To demonstrate how the credit segment compiler handles predictions with complex dependencies,  
we implement an arc-eager transition-based dependency parser~\cite{nivre03parsing}.
At each state, it  takes one of the four actions $\{Shift, Reduce, Left, Right\}$ based on
a simple neural network with one hidden layer of size 5 and generates a dependency parse to a sentence in the end.
The rollin policy is the current (learned) policy.
The probability of executing the reference policy (dynamic oracle)~\cite{goldberg13oracles}
for rollout decreases over each round.
We compare our model with two recent greedy transition-based parsers implemented by the original authors,
the dynamic oracle parser (\dn{})~\cite{goldberg13oracles}
and the Stanford neural network parser (\nn{})~\cite{danqi14nndep}.
We also present the best results in CoNLL-X and the best published results for CTB and PTB.
The performances are evaluated by unlabeled attachment scores (UAS). Punctuation is excluded.

Table~\ref{tab:resultdep} shows the results.  Our implementation with
only \~{}300 lines of C++ code is competitive with $\dn{}$ and
$\nn{}$, which are specifically designed for parsing. Remarkably, our
system achieves strong performance on CoNLL-X without tuning any
hyper-parameters, even beating heavily tuned systems participating in the challenge on one dataset.
The best system to date on PTB~\cite{andor16} uses a global
normalization, more complex neural network layers and k-best POS
tags. Similarly, the best system for CTB~\cite{dyer15} uses stack
LSTM architectures tailored for dependency parsing.

\begin{table*}[!t]
\centering
\begin{tabular}{l|c@{\enskip}c@{\enskip}c@{\enskip}c@{\enskip}c@{\enskip}c@{\enskip}c@{\enskip}c@{\enskip}c@{\enskip}c|c@{\enskip}c}
\toprule
Parser & \textsc{Ar} & \textsc{Bu} & \textsc{Ch} & \textsc{Cz$^{+}$} &
\textsc{Da} & \textsc{Du$^{+}$} & \textsc{Ja$^{+}$} & \textsc{Po$^{+}$} & 
\textsc{Sl$^{+}$} & \textsc{Sw} & \textsc{PTB} & \textsc{CTB} \\
\midrule
\dn{} & 75.3 & 89.8 & 88.7 & 81.5     & 87.9 & 74.2 & \bf{92.1} & 88.9 & 78.5 & 88.9 & 90.3 & 80.0 \\
\nn{} & 67.4$^*$ & 88.1 & 87.3 & 78.2 & 83.0 & 75.3 & 89.5 & 83.2$^*$ & 63.6$^*$ & 85.7 & 91.8$^{\#}$ & 83.9$^{\#}$\\
\our{} &\bf{78.2}& \bf{92.0} & \bf{89.8} &  \bf{84.8} & \bf{89.8} & \bf{79.2} & 91.8 & \bf{90.6} & \bf{82.2} & \bf{89.7} & 
\bf{91.9} & \bf{85.1} \\
\midrule
{\sc Best} & 79.3 & 92.0 & 93.2 & 87.30& 90.6 & 83.6& 93.2 & 91.4 & 83.2 & 
89.5 & 94.4$^{\#}$ & 87.2$^{\#}$\\
\bottomrule
\end{tabular}
\caption{UAS on PTB, CTB and CoNLL-X. Best: the best known result in
  CoNLL-X or the best published results (CTB, PTB) using arbitrary
  features and resources.  See details and additional results in text
  and in the appendix.\protect\footnotemark }
\label{tab:resultdep}
\end{table*}

\footnotetext{($^*$) \nn{} makes assumptions about the structure of languages and hence obtains substantially worse performance on languages with multi-root trees.   
($^{+}$) Languages contains more than 1\% non-projective arcs, where a 
transition-based parser (e.g. \our)
likely underperforms graph-based parser ({\sf Best}) due to the model assumptions.
($^{\#}$) Numbers reported in the published papers~\cite{danqi14nndep,dyer15,andor16}.}


\subsection{Empirical evaluation of optimizations}
\label{sec:pathcollapse}

In Section~\ref{sec:comp}, we discussed two approaches for
computational improvements. Memoization avoids re-predicting on the
same input multiple times while path collapse stops rollouts at a
particular point in time.  The effect of the different optimizations
depends greatly on the underlying learning algorithm.  For example,
DAgger does not do rollouts at all, so no efficiency is gained by
either optimization.\footnote{Training speed is only degraded by about 0.5\% with optimizations on, demonstrating negligible overhead.} The affected algorithms are LOLS (with mixed
rollouts) and Searn.

\begin{figure}[t]

\begin{center}
	\begin{tabular}{@{}p{0.6\linewidth}@{}p{0.38\linewidth}}
	\begin{tabular}{|l|c|c|c|c|}
\hline
& \multicolumn{2}{c|}{\textbf{             NER}} & \multicolumn{2}{c|}{\textbf{  POS}} \\ \hline
                                 &   LOLS    &       Searn    &     
								 LOLS     &      Searn \\								 
								 \hline
\textbf{No Opts               }  &   96s          &       123s     &     3739s         &      4255s \\ 
\textbf{Mem.             }  &   75s          &       85s      &     1142s         &      1215s \\
\textbf{Col.@4+Mem.}  &   71s          &       75s      &     1059s         &      1104s \\ 
\textbf{Col.@2+Mem.}  &   69s          &       71s      &     1038s         &      1074s \\ \hline
\end{tabular}&
\vspace{-0.6in}
\includegraphics[width=\linewidth]{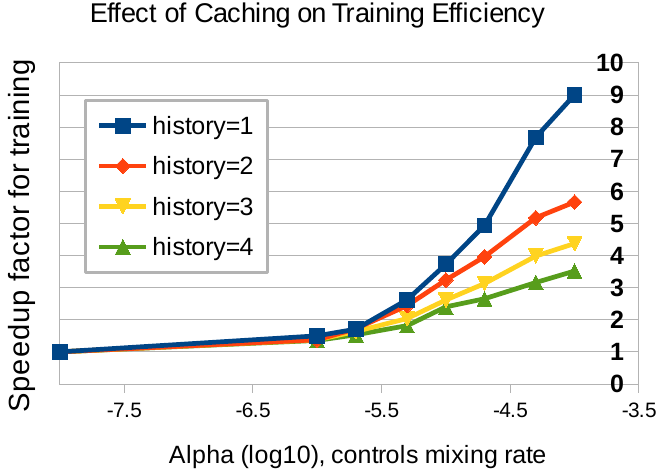} \\
\end{tabular}
\end{center}
\caption{The table on the left shows the effect of Collapse (Col) and Memorization (Mem.).
The figure on the right shows the speed-up obtained for different historical 
lengths and mixing rate of rollout policy.  Large $\alpha$ corresponds to more 
prediction required when training the model.  
}
\label{fig:opts}
\end{figure}

Figure \ref{fig:opts} shows the effect of these optimizations on the \emph{best} 
NER and POS systems we trained without using external resources. In the left table,
we can see that memoization alone reduces overall training runtime by 
about $25\%$ on NER and about $70\%$ on POS, essentially because the overhead 
for the classifier on POS tagging is so much higher (45 labels versus 9). When 
rollouts are terminated early, the speed increases are much more modest, 
essentially because memoization is already accounting for much of these gains. 
In all cases, the final performance of the predictors is within statistical 
significance of each other (p-value of 0.95, paired sign test), \emph{except} 
for Collapse@2+Memoization on NER, where the performance decrease is only 
insignificant at the 0.90 level. The right figure demonstrates that when $\alpha$
increases, more prediction is required during the training time, and 
the speedup increases from a factor of 1 (no change) to a factor of as much as 9.
However, as the history length increases, the speedup is more modest due to 
low cache hits.

\section{Related Work}
\label{sec:pprog}
Several algorithms are similar to learning to search approaches,
including the incremental structured perceptron
\cite{collins04incremental,huang12structured},
HC-Search~\cite{doppa12oss,doppa14hcsearch}, and others
\cite{daume05laso,ratliff07boosting,syed11reduction,xu07beam,xu07planning}.
Some fit this framework.

Probabilistic programming~\cite{gordon14probprog} has been an active
area of research.  These approaches have a different goal: Providing a
flexible framework for specifying graphical models and performing
inference in those models.  The credit assignment compiler instead
allows a developer to learn to make coherent decisions for joint
prediction (``learning to search'').  We also differ by not designing
a new programming language. Instead, we have a two-function library
which makes adoption and integration into existing code bases much
easier.  

The closest work to ours is
Factorie~\cite{mccallum09factorie}. Factorie is essentially an
embedded language for writing factor graphs compiled into Scala to run
efficiently.\footnote{Factorie-based implementations of simple tasks
  are still less efficient than systems like \system{CRF SGD}. }
Factorie
acts more like a library than a language although it's abstraction is still factor graph distributions.
Similarly, Infer.NET~\citep{minkainfer}, Markov Logic Networks (MNLs)
~\citep{richardson2006markov}, and Probabilistic Soft Logic
(PSL)~\citep{kimmig2012short} concisely construct and use
probabilistic graphical models.  BLOG~\cite{milch2007blog} falls in
the same category, though with a very different focus. Similarly,
Dyna~\cite{eisner2005compiling} is a related declarative language for
specifying probabilistic dynamic programs, and Saul~\cite{Parisa15} is
a declarative language embedded in Scala that deals with joint
prediction via integer linear programming.
All of these examples have picked particular aspects of the
probabilistic modeling framework to focus on.  Beyond these examples,
there are several approaches that essentially ``reinvent'' an existing
programming language to support probabilistic reasoning at the first
order level. IBAL~\cite{pfeffer2001ibal} derives from O'Caml;
Church~\cite{goodman2008church} derives from LISP. IBAL uses a (highly
optimized) form of variable elimination for inference that takes
strong advantage of the structure of the program; Church uses MCMC
techniques, coupled with a different type of structural reasoning to
improve efficiency.  

\bibliographystyle{abbrv}

\bibliography{bibfile}



\newpage
\appendix
\section{Example TDOLR programs}

In this section, a few other TDOLR programs which illustrate the
ease and flexibility of programming. 

Algorithm~\ref{alg:sdetect} is for a sequential \emph{detection} task
where the goal is to detect whether or not a sequence contains some
rare element.  This illustrates outputs of lengths other than the
number of examples, explicit loss functions.

\newalgorithm{sdetect}%
  {\FUN{Sequential\_Detection\_RUN}(\VAR{examples} as $X$, \VAR{false\_negative\_loss})}%
  {
    \STATE Let \VAR{max\_value} = 1
    \FOR{\VAR{$i$} $=\CON{1}$ \TO \FUN{len}(\VAR{examples})}
      \SETST{max\_value}{\FUN{max}(\VAR{max\_value}, \VAR{examples}{[}\VAR{i}{]}\texttt{.}\VAR{label})}
    \ENDFOR

    \STATE Let \VAR{max\_prediction} = 1

    \FOR{\VAR{$i$} $=\CON{1}$ \TO \FUN{len}(\VAR{examples})}
      \SETST{max\_prediction}{\FUN{max}(\VAR{max\_prediction}, \FUN{predict}($x$=\VAR{examples}{[}\VAR{i}{]}, $y$=\VAR{examples}{[}\VAR{i}{]}\texttt{.}\VAR{label}))} \COMMENT{maintain max}
    \ENDFOR

    \IF{\VAR{max\_label} > \VAR{max\_prediction}} 
       \STATE \FUN{loss}(\VAR{false\_negative\_loss}) \COMMENT{The loss is asymmetric}
    \ELSE 
       \IF{\VAR{max\_label} < \VAR{max\_prediction}}
           \STATE \FUN{loss}(1)
       \ELSE
           \STATE \FUN{loss}(0)
       \ENDIF
    \ENDIF

    \IF{\VAR{output}\texttt{.}\VAR{good}}
        \STATE \VAR{output} \texttt{<<} \VAR{max\_prediction}\COMMENT{if we should generate output, append our prediction}
    \ENDIF
  }

In Algorithm~\ref{alg:depparsing}, we show an implementation of a shift-reduce dependency parser for natural language. We discuss each subcomponent below.  
The detailed introduction to dependency parsing is provided in the next section.  
\begin{itemize}
\item \textsc{GetValidAction} returns valid actions that can be taken by the current configuration.
\item \textsc{GetFeat} extracts features based on the current configuration.   
A detailed list of our features is in the supplementary material.
\item \textsc{GetGoldAction} implements the dynamic oracle described in \cite{goldberg13oracles}. 
The dynamic oracle returns the optimal action in any state that leads to a reachable end state with the minimal loss.
\item \textsc{Predict} is a library call implemented in the L2S system.  
Given training samples,  L2S can learn the policy automatically.
In the test phase, it returns a predicted action leading to an end state with small structured loss.
\item \textsc{Trans} implements the hybrid-arc transition system described above.
\item \textsc{Loss} 
returns the number of words whose parents are wrong. It has no effect in the test phase.
\end{itemize}

\newalgorithm{depparsing}%
  {\FUN{RunParser}(\VAR{sentence} as $X$)}%
  {
  \SETST{{stack $S$}}{\{\bf Root\}}
    \SETST{{buffer $B$}}{[words in sentence]}
    \SETST{{arcs $A$}}{$\emptyset$}
	\WHILE{{ $B\neq \emptyset$ or $|S| > 1$}}
	  \SETST{ValidActs}{\FUN{GetValidActions}($S, B$)}
	  \SETST{features}{\FUN{GetFeat}($S, B, A$)}
	  \SETST{ref}{\FUN{GetGoldAction}($S, B$)}
	  \SETST{action}{\FUN{predict}({$x$=features, $y$=ref, ValidActs})}
	  \SETST{$S, B, A$}{\FUN{Trans}($S, B , A$, action)}
    \ENDWHILE
    \STATE \FUN{loss}($A[w]$ $\neq$ $A^*[w]$, $\forall w \in$ sentence)
    \STATE \textbf{return} output
  }

We show that this parser achieves strong results across ten languages from the CoNLL-X 
challenge and performs well on two standard evaluation data sets, and requires only about $300$ lines of readable C++ code.

Finally, Algorithm \ref{alg:entrel} provides an implementation for
jointly assigning types to name entities in a sentence and recognizing relations between 
them~\cite{RothYi07}. Besides features used for predicting entity and relation 
types. We also consider constraints that ensure the entity-type 
assignments and relation-type assignments are compatible with each other. 
For example, the first argument of the {work\_for} relation need to be tagged 
as person, and the second argument has to be an organization.

\newalgorithm{entrel}%
  {\FUN{Entity\_Relation\_RUN}(\VAR{sent} as $X$)}%
  {
    \SETST{\VAR{output}}{\FUN{Initialize\_structure}()}
    \SETST{\VAR{K}}{\FUN{Number\_of\_entities}(\VAR{sent})}
    \FOR{{\VAR{n}} = \CON{1} to \VAR{K}}
      \SETST{ref}{\VAR{sent\texttt{.}entity\_type}[\VAR{n}]}
      \SETST{\VAR{output}\texttt{.}entity\_type[n]}{\FUN{predict}(\texttt{\footnotesize $x$=}\VAR{sent}.entity[i], \texttt{\footnotesize $y$=}\VAR{ref}, \texttt{\footnotesize tag=}\VAR{n}}
 \STATE \FUN{loss}(\VAR{output}\texttt{.}entity\_type[n] $\neq$ \VAR{sent}\texttt{.}entity\_type[n]\texttt{.}true\_label)
    \ENDFOR
    \FOR{{\VAR{n}} = \CON{1} to \VAR{K}-1}
    \FOR{{\VAR{m}} = \CON{n+1} to \VAR{K})}
      \SETST{ref}{\VAR{sent}\texttt{.}relation\_type[\VAR{n},\VAR{m}]\texttt{.}{true\_label}}
      \SETST{valid\_relations}{\FUN{Find\_valid\_relations}(\VAR{output}\texttt{.}entity\_type[\VAR{n}], \VAR{output}\texttt{.}entity\_type[\VAR{m}])}
      \SETST{\VAR{output\texttt{.}relation\_type}}{\FUN{predict}(\texttt{\footnotesize $x$=}\VAR{sent}.relation[n,m], \texttt{\footnotesize $y$=}\VAR{ref}, \texttt{\footnotesize tag=}\VAR{K*(n+1)+m}, \texttt{\footnotesize valid\_labels=}\VAR{valid\_relations}, \texttt{\footnotesize condition=}[\VAR{n},\VAR{m}])}
 \STATE \FUN{loss}(\VAR{output}\texttt{.}relation\_type[n,m] $\neq$ \VAR{sent}\texttt{.}entity\_type[n,m]\texttt{.}true\_label )
\ENDFOR
\ENDFOR
    \STATE \textbf{return} \VAR{output}
  }

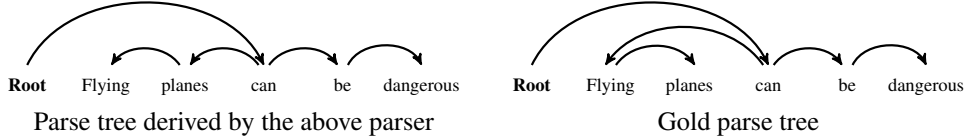
\begin{figure*}
	\resizebox{\textwidth}{!}{
	\begin{tabular}{@{ }l|r@{ }l@{ }l@{}}
		\multirow{2}{*}{Action} & \multicolumn{3}{c}{Configuration}\\
		& {\bf S}tack & {\bf B}uffer & {\bf A}rcs \\
		\hline
		& [{\bf Root}] & [Flying planes can be dangerous]  & \{\}\\
		\textsc{Shift} &[{\bf Root} Flying] & [planes can be dangerous] & \{\}\\		
		\textsc{Reduce-left} & [{\bf Root}] &[planes can be dangerous] & \{(planes, Flying)\}\\
		\textsc{Shift} &[{\bf Root} planes] & [can be dangerous] & \{(planes, Flying)\}\\
		\textsc{Reduce-left} & [{\bf Root}] &[can be dangerous] & \{(planes, Flying), (can, planes)\}\\
		\textsc{Shift} & [{\bf Root} can] &[be dangerous] & \{(planes, Flying), (can, planes)\}\\
		\textsc{Shift} & [{\bf Root} can be] &[dangerous] & \{(planes, Flying), (can, planes)\}\\		
		\textsc{Shift} & [{\bf Root} can be dangerous] &[] & \{(planes, Flying), (can, planes)\}\\		
		\textsc{Reduce-Right} & [{\bf Root} can be] &[] & \{(planes, Flying), (can, planes), (be, dangerous)\}\\		
		\textsc{Reduce-Right} & [{\bf Root} can] &[] & \{(planes, Flying), (can, planes), (be, dangerous), (can, be)\}\\		
		\textsc{Reduce-Right} & [{\bf Root}] &[] & \{(planes, Flying), (can, planes), (be, 
		dangerous), (can, be), ({\bf Root}, can)\}\\		
	\end{tabular}
	}
\\
\begin{tabular}{cc}		
\resizebox{0.45\textwidth}{!}{
\begin{tikzpicture}[start chain,->,>=stealth',node 
  distance=0.5cm,very thick,every node/.style={anchor=base}]
  \node   (A) at (0,1)   {\bf Root};
  \node   (B) at (1.5,1) {Flying};
  \node   (C) at (3,1) {planes};
  \node   (D) at (4.5,1)  {can};
  \node   (E) at (6,1)  {be};
  \node   (F) at (7.5,1)  {dangerous};
  \path (0,1.5)    edge [bend left=64]   (4.5,1.5)
	    (4.4,1.5) edge [bend right=55]     (3.1,1.5)
	    (2.9,1.5) edge [bend right=55]     (1.6,1.5)		
        (4.6,1.5) edge [bend left=55]     (5.9,1.5)
        (6.1,1.5)   edge [bend left=64]      (7.5,1.5);
\end{tikzpicture}
}&
\resizebox{0.45\textwidth}{!}{
\begin{tikzpicture}[start chain,->,>=stealth',node 
  distance=0.5cm,very thick,every node/.style={anchor=base}]
  \node   (A) at (0,1)   {\bf  Root};
  \node   (B) at (1.5,1) {Flying};
  \node   (C) at (3,1) {planes};
  \node   (D) at (4.5,1)  {can};
  \node   (E) at (6,1)  {be};
  \node   (F) at (7.5,1)  {dangerous};
  \path (0,1.5)    edge [bend left=64]   (4.5,1.5)
	    (1.6,1.5) edge [bend left=55]     (3.1,1.5)
	    (4.4,1.5) edge [bend right=55]     (1.4,1.5)		
        (4.6,1.5) edge [bend left=55]     (5.9,1.5)
        (6.1,1.5)   edge [bend left=64]      (7.5,1.5);
\end{tikzpicture}
}\\
Parse tree derived by the above parser & Gold parse tree
\end{tabular}
	\caption{An illustrative example of an arc-hybrid transition parser.  
	The above table show the actions taken and the intermediate configurations generated by 
	a parser.  The parse tree derived by the parser is in the bottom left, and 
	the gold parse tree is the bottom right.  
	The distance between these two trees is 2.	}

	\label{tab:dep}
\end{figure*}


\section{Dependency Parsing}
In the following, we provide a brief overview of transition-based dependency  parsing.    
A transition-based dependency parser takes a sequence of actions and parses a 
sentence from left to right by maintaining a \emph{stack} $S$, a 
\emph{buffer} $B$, and a set of \emph{dependency arcs} $A$.  
The stack maintains partial parses, the buffer stores the 
words to be parsed, and $A$ keeps the arcs that have been generated so far.  
The configuration of the parser at each stage can be defined by a triple
$(S, B, A)$.
For the ease of notation, we use $w_p$ to represent the leftmost word in the buffer 
and use $s_1$ and $s_2$ to denote the top and the second top 
words in the stack. A dependency arc $(w_h,w_m)$ is a directed edge that indicates word $w_h$
is the parent of word $w_m$. When the parser terminates, the arcs in $A$ form a projective dependency tree.  
We assume that each word only has one parent in the derived dependency parse tree, and use $A[w_m]$ to denote the parent of word $w_m$.
For labeled dependency parsing, we further assign a tag to each arc representing the dependency type between the head and the modifier.  
For simplicity, we assume an unlabeled parser in the following description.
The extension from an unlabeled parser to a labeled parser is straightforward, and is discussed at the end of this section.

\newalgorithm{Trans}%
{\FUN{Trans}($S$, $B$, $A$, action)}%
  {
	\STATE{Let $w_p$ be the leftmost element in $B$}
  	\IF{action = \textsc{Shift}}
  	    \STATE{$S$.push($w_p$)}
  	    \STATE{remove $w_p$ from $B$}
	\ELSIF{action= \textsc{Reduce-Left}}
		\SETST{top}{$S$.pop()}
		\SETST{$A$}{$A \cup$ ($w_p$,top)}
	\ELSIF{action = \textsc{Reduce-Right}}
		\SETST{top}{$S$.pop()}
		\SETST{$A$}{$A \cup$ ($S$.top(), top)}
	\ENDIF
	\STATE{\textbf{return} $S, B, A$}
  }

In the following, we describe an arc-hybrid transition system due to its 
simplicity. The arc-eager system used in the experiments share the same spirit.  
In the initial configuration, the buffer $B$ contains all the words in the sentence,
a dummy root node is pushed in the stack $S$, and the set of arcs $A$ is empty. 
The root node cannot be popped out at anytime during parsing.
The system then takes a sequence of actions until the buffer is empty  and the 
stack contains only the root node (i.e., $|B|=0$ and $S=\{\textbf{Root}\}$). When the process terminates, a parse tree is derived.
At each state, the system can take one of 
the following actions:
\begin{enumerate}
	\item \textsc{Shift}: push $w_p$ to $S$ and move $p$ to the next word. (Valid when $|B|>0$).
\item \textsc{Reduce-left}: add an arc ($w_p$, $s_1$)
	to $A$ and pop $s_1$. (Valid when $|B| > 0$ and $|S| > 1$).
\item \textsc{Reduce-right}: add an arc ($s_2$, $s_1$) 
	to $A$ and pop $s_1$. (Valid when  $|S| > 1$).
\end{enumerate}
Algorithm \ref{alg:Trans} shows the execution of these actions during parsing, and
Figure \ref{tab:dep} demonstrates an example of transition-based dependency parsing.  

\section{Additional Experiment Results}

\subsection{Sequential Tagging}
\begin{figure}
\begin{tabular}{@{}c@{ }c@{}}
\includegraphics[width=0.5\textwidth,clip=true,trim=24 25 38 20]{POS0.pdf}&
\includegraphics[width=0.5\textwidth,clip=true,trim=24 25 38 20]{NER0.pdf}\\
\includegraphics[width=0.5\textwidth,clip=true,trim=24 25 38 20]{POS1.pdf}&
\includegraphics[width=0.5\textwidth,clip=true,trim=24 25 38 20]{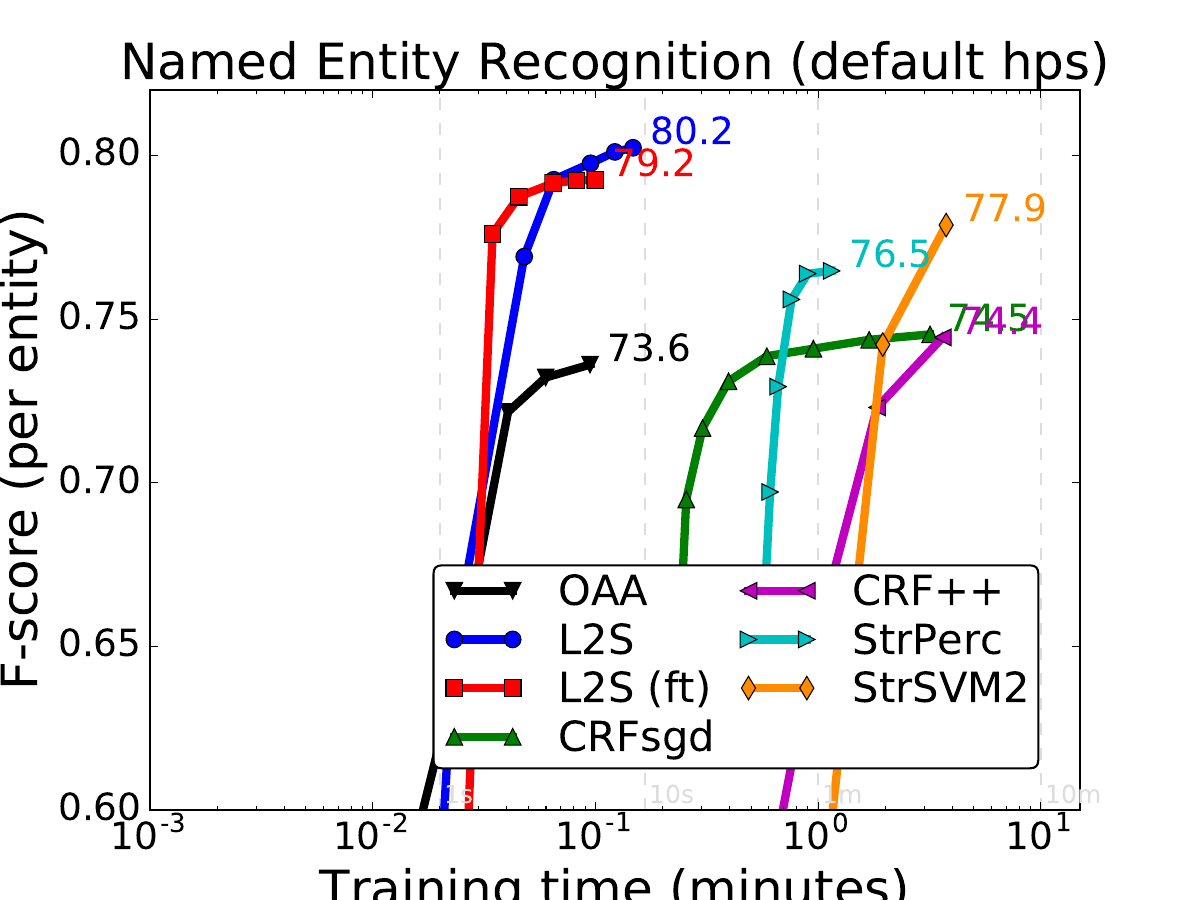}
\end{tabular}
\caption{Training time versus evaluation accuracy for part of speech tagging (left) and named entity recognition (right). X-axis is in log scale. Different points correspond to different termination criteria for training. Top figures use hyperparameters that were tuned (for accuracy) on the holdout data. (Note: lines are curved due to log scale x-axis.)}
\label{fig:results-details}
\end{figure}

In Figure~\ref{fig:results-details}, we enlarge the figures in 
\ref{fig:results} and provide the results of  NER with default parameters.

\subsection{Dependency Parsing}

\begin{table*}[!t]
\centering
\begin{tabular}{l|c@{\enskip}c@{\enskip}c@{\enskip}c@{\enskip}c@{\enskip}c@{\enskip}c@{\enskip}c@{\enskip}c@{\enskip}c|c@{\enskip}c}
\toprule
Parser & \textsc{Ar} & \textsc{Bu} & \textsc{Ch} & \textsc{Cz$^{+}$} &
\textsc{Da} & \textsc{Du$^{+}$} & \textsc{Ja$^{+}$} & \textsc{Po$^{+}$} & 
\textsc{Sl$^{+}$} & \textsc{Sw} & \textsc{PTB} & \textsc{CTB} \\
\midrule
& \multicolumn{11}{c}{UAS} & \\
\midrule
\dn{} & 75.3 & 89.8 & 88.7 & 81.5     & 87.9 & 74.2     & 92.1 & 88.9 & 78.5 & 88.9 & 90.3 & 80.0 \\
\nn{} & 67.4$^*$ & 88.1 & 87.3 & 78.2 & 83.0 & 75.3 & 89.5 & 83.2$^*$ & 63.6$^*$ & 85.7 & 91.8$^{\#}$ & 83.9$^{\#}$\\
\our{}$^{O}$ &75.3 & 89.5 & 87.4 & 81.1     & 86.0 & 75.3 & 90.4 & 88.4 & 78.5 & 89.9 & 91.9  & 85.1\\
\our{} &78.2& 92.0 & 89.8 & 84.8     & 89.8 & 79.2 & 91.8 & 90.6 & 82.2 
& 89.7 & 91.9 & 85.1 \\
{\sf Best} & 79.3 & 92.0 & 93.2 & 87.30& 90.6 & 83.6& 93.2 & 91.4 & 83.2 & 89.5 & 94.4$^{\#}$ & 87.2$^{\#}$\\
\midrule
& \multicolumn{11}{c}{LAS} & \\
\midrule
\dn{}  & 64.3 & 85.0 & 84.6 &  74.1   & 82.5 &  70.3    & 90.6 & 85.0 & 68.5 & 83.5 & 88.1 & 78.8\\
\nn{} & 51.7$^*$  & 84.0 & 82.7 & 77.4 & 72.0 & 89.1 & 87.4 $^*$ & 77.9 $^*$ & 51.1$^*$ & 80.1 & 89.6$^{\#}$ & 82.4$^{\#}$\\
\our{}$^{O}$ & 65.1 & 85.0 & 80.8 & 74.5    & 81.0 & 72.1 & 88.4 & 84.4 & 69.4 & 85.2 &  89.7  & 83.6\\
\our{}& 68.2 & 88.2 & 87.1 &  79.6      & 84.9 & 75.8 & 89.7 & 87.8 & 74.0 & 84.9 & 89.7 & 83.6\\
{\sf Best} & 66.9 & 87.6 & 90.0 & 80.2& 84.8 & 79.2& 91.7 & 87.6 & 73.4$^*$ & 84.6 & 92.55$^{\#}$ & 85.7$^{\#}$ \\
\bottomrule
\end{tabular}
\caption{Accuracy on PTB, CTB and CoNLL-X. Best: best results from the shared 
task. Best: the best results reported in CoNLL-X (may come from different 
participants) and the best published results (CTB,PTB). 
\our{}$^O$, \dn{}, \nn{} use only features generated by word and POS tags, 
while \our{} and the models in CoNLL-X use additional morphologic features. 
\our{} models use Brown cluster for PTB, $\dn{}$ and the {\sf Best} use word 
embedding features generated from unsupervised text corpus with billion words.  
The {\sf Best} models also used word embedding features for $CTB$.\protect\footnotemark
}
\label{tab:result-depdetails}
\end{table*}

\footnotetext{($^*$) \nn{} makes assumptions about the structure of languages and hence obtains substantially worse performance on languages with multi-root trees.   
($^{+}$) Languages contains more than 1\% non-projective arcs, where a 
transition-based parser (e.g. \our)
likely underperforms graph-based parser ({\sf Best}) due to the model assumptions.
($^{\#}$) Numbers reported in the published papers~\cite{danqi14nndep,dyer15,andor16}.}

\begin{table*}[!t]
\centering
\begin{tabular}{l|c@{\enskip}c@{\enskip}c@{\enskip}c@{\enskip}c@{\enskip}c@{\enskip}c@{\enskip}c@{\enskip}c@{\enskip}c|c@{\enskip}c}
\toprule
Parser & \textsc{Ar} & \textsc{Bu} & \textsc{Ch} & \textsc{Cz$^{+}$} &
\textsc{Da} & \textsc{Du$^{+}$} & \textsc{Ja$^{+}$} & \textsc{Po$^{+}$} & 
\textsc{Sl$^{+}$} & \textsc{Sw} & \textsc{PTB} & \textsc{CTB} \\
\midrule
\dn{} &     &  & & r  & & r &r & r & r & & T & T \\
\nn{} & r'  &  & & r  & & r &r  & r' & r' &  & T & T\\
\our{}$^{O}$& & &  &  r  &  & r & r &r &r &  & T  & T\\
\our{} & M &M & M & rM     & M & rM & rM & rM & rM & M & ET & T  \\
{\sf Best} & TM & TM & TM & TM& TM & TM & TM & TM & TM & TM & ET &  ET  \\
\midrule

\end{tabular}
\caption{
T: tuned hyper-parameters, M: use morphological features, E: use word 
embedding or word clustering, r: language structure assumption that may 
degrades the performance (the nature of transition-based model), r': strong 
language structure assumption (only one head) that severely degrades the performance.   
Accuracy on PTB, CTB and CoNLL-X. Best: best results from the shared }
\label{tab:depdiff}
\end{table*}

Table \ref{tab:result-depdetails} show the complete experiment results for 
dependency parsing. The system is evaluated on both unlabeled attachment 
score (UAS) and labeled attachment score. Again, conducting fair comparisons 
across different systems is difficult because different systems use 
different sets of features and different assumptions about the structure of languages.   
Table \ref{tab:depdiff} summarizes the differences.

\section{Experiment details}

\subsection{Datasets and Tasks}

\begin{table}
\begin{tabular}{|l|rrrrr|rr|rr|}
\hline
&
\multicolumn{5}{c|}{\textbf{Training}} &
\multicolumn{2}{c|}{\textbf{Holdout}} &
\multicolumn{2}{c|}{\textbf{Test}} \\
      & Sents &  Toks  & Labels & Features & Unique Fts & Sents & Toks & Sents & Toks \\
\hline
\textbf{POS} &  38k &  912k & 45 & 13,685k &  629k &  5.5k  &     132k  &    5.5k  &      130k \\
\textbf{NER} &  15k &  205k &  7 &  8,592k &  347k &  3.5k  &      52k  &    3.6k  &       47k \\
\hline
\end{tabular}
\caption{Basic statistics about the data sets used for part of speech (POS) tagging and named entity recognition (NER).}
\label{tab:datasets}
\vspace{-1em}
\end{table}


\begin{figure}
\begin{tabular}{|l|l|}
\hline
\textbf{POS} &
\begin{tabular}{c@{ }c@{ }c@{ }c@{ }c@{ }c@{ }c@{ }c@{ }c@{ }c@{ }c@{ }c@{ }c@{ }c@{ }c@{ }c}
\wordtag{NNP} & \wordtag{NNP} & \wordtag{,} & \wordtag{CD} & \wordtag{NNS} & \wordtag{JJ} & \wordtag{,} & \wordtag{MD} & \wordtag{VB} & \wordtag{DT} & \wordtag{NN} & \wordtag{IN} & \wordtag{DT} & \wordtag{JJ} & \wordtag{NN} & \\
Pierre & Vinken & , & 61 & years & old & , & will & join & the & board & as & a & nonexecutive & director & \dots
\end{tabular} \\
\hline
\textbf{NER} 
&
\overbracetxt{Germany}{LOC} 's rep to the \overbracetxt{European Union}{ORG} 's committee \overbracetxt{Werner Zwingmann}{PER} said \dots \\
\hline
\end{tabular}
\caption{Example inputs (below, black) and desired outputs (above, blue) for 
part of speech tagging task, named entity recognition task, and 
entity-relation recognition task. }
\label{fig:examples}
\end{figure}
We conduct experiments based on two variants of the sequence labeling problem 
(Algorithm~\ref{alg:sequential}) The first is a pure sequence labeling 
problem: Part of Speech tagging based on data from the Wall Street Journal 
portion of the Penn Treebank. The second is a sequence \emph{chunking} 
problem: named entity recognition using data from the CoNLL 2003 dataset.
See Figure~\ref{fig:examples} for example inputs and outputs for these tasks.%

Part of speech tagging for English is based on the Penn Treebank tagset that includes $45$ discrete labels.
The accuracy reported represents number of tokens tagged correctly. This is a \emph{pure} sequence labeling task. 
Named entity recognition for English is based on the CoNLL 2003 dataset that includes four entity types: Person, Organization, Location and Miscellaneous. We use the standard evaluation metric to report performance as macro-averaged F-measure. 
In order to cast this \emph{chunking} task as a sequence labeling task, we use the standard Begin-In-Out (BIO) encoding, though some results suggest other encodings may be preferable \cite{ratinov09design} (we tried BILOU and our accuracies decreased).
The example sentence from Figure~\ref{fig:examples} in this encoding is:
\begin{center}
\begin{tabular}{c@{ }c@{ }c@{ }c@{ }c@{ }c@{ }c@{ }c@{ }c@{ }c@{ }c}
\overbracetxt{Germany}{LOC} & 's & rep & to & the & \overbracetxt{European Union}{ORG} & 's & committee & \overbracetxt{Werner Zwingmann}{PER} & said & \dots \\
\wordtag{B-LOC} & \wordtag{O} & \wordtag{O} & \wordtag{O} & \wordtag{O} & \wordtag{~~~B-ORG~~~~~~~I-ORG} & \wordtag{O} & \wordtag{O} & \wordtag{B-PER~~~~~~~~~~I-PER~~~~~} & \wordtag{O} &
\end{tabular}
\end{center}

Dependency parser is test on the English Penn Treebank (PTB) and 
the CoNLL-X datasets for 9 other languages, including
Arabic, Bulgarian, Chinese, Danish, Dutch, Japanese, Portuguese, Slovene and Swedish.
For PTB, we convert the constituency trees to dependencies by the Stanford 
parser 3.3.0.
We follow the standard split: sections 2 to 21 for training,
section 22 for development and section 23 for testing.
The POS tags in the evaluation data is assigned by the Stanford POS tagger, which has an accuracy of 97.2\% on the PTB test set.
For CoNLL-X, we use the given train/test splits and reserve the last 10\% of training data for development if needed.
The gold POS tags given in the CoNLL-X datasets are used.
The CTB is prepared following the instructions in \cite{danqi14nndep}.

\subsection{Methodology}

Comparing different systems is challenging because one wishes to hold constant as many variables as possible. In particular, we want to control for both \textbf{features} and \textbf{hyperparameters}. In general, if a methodological decision cannot be made ``fairly,'' we made it in favor of competing approaches.

To control for \textbf{features}, for the two sequential tagging tasks (POS 
and NER), we use the built-in \emph{feature template} approach of 
\system{CRF++}\ (duplicated in \system{CRF SGD}) to generate features. The 
other approaches (\system{Structured SVM}, \system{VW Search}\ and \system{VW 
Classification}) all use the features generated (offline) by \system{CRF++}. 
For each task, we tested six feature templates and picked the one with best 
development performance using \system{CRF++}. The templates included 
neighboring words and, in the case of NER, neighboring POS tags. 
\emph{However,} because \system{VW Search}\ is also able to generate features 
from its own templates, we also provide results for \textbf{\system{VW Search\ 
(own fts)}} in which it uses its own, internal, feature template generation, 
which were tuned to maximize it's holdout performance on the most 
time-consuming run (4 passes) and include neighboring words (and POS tags, for 
NER) and word prefixes/suffixes.\footnote{The exact templates used are 
provided in the supplementary materials.} In all cases we use \emph{first 
order Markov dependencies,} which lessens the speed advantage of search based 
structured prediction.

To control for \textbf{hyperparameters}, we first separated each system's hyperparameters into two sets: (1) those that affect termination condition and (2) those that otherwise affect model performance. When available, we tune hyperparameters for (a) learning rate and (b) regularization strength\footnote{Precise details of hyperparameters tuned and their ranges is in the supplementary materials.}. Additionally, we vary the termination conditions to sweep across different amounts of time spent training. For each termination condition, we can compute results using either the \textbf{default hyperparameters} or the \textbf{tuned hyperparameters} that achieved best performance on holdout data. We report both conditions to give a sense of how sensitive each approach is to the setting of hyperparameters (the amount of hyperparameter tuning directly affects effective training time).

One final confounding issue is that of \textbf{parallelization}. Of the baseline approaches, only \system{CRF++}\ supports parallelization via multiple threads at training time. In our reported results, \system{CRF++}'s time is the total CPU time (i.e., effectively using only one thread). Experimentally, we found that wall clock time could be decreased by a factor of $1.8$ by using $2$ threads, a factor of $3$ using $4$ threads, and a (plateaued) factor of $4$ using $8$ threads. This should be kept in mind when interpreting results. DEMI-DCD (for structured SVMs) also \emph{must} use multiple threads. To be as fair as possible, we used $2$ threads. Likewise, it can be sped up more using more threads \cite{chang13svm}. VW (Search and Classification) can also easily be parallelized using AllReduce \cite{agarwal11allreduce}. We do not conduct experiments with this option here because none of our training times warranted parallelization (a few minutes to train, max).

For dependency parsing, we fixed the hyper-parameters when test on CoNLL-X.   
For CTB and PTB, we tune the size of beam in beam search and the history 
length of predictions.  For PTB, we further use dictionary features from Brown 
cluster.   

\subsection{Hardware Used}

All timing results were obtained on the same machine with the
following configuration. Nothing else was run on this machine
concurrently:

\begin{verbatim}  
%   2 * Intel(R) Core(TM)2 Duo CPU E8500 @ 3.16GHz
  6144 KB cache
  8 GB RAM, 4 GB Swap
  Red Hat Enterprise Linux Workstation release 6.5 (Santiago)
  Linux 2.6.32-431.17.1.el6.x86_64 #1 SMP
    from Fri Apr 11 17:27:00 EDT 2014 x86_64 x86_64 x86_64 GNU/Linux
\end{verbatim}

\subsection{Software Used}

The precise software versions used for comparison are:

\begin{compactitem}
\item[\system{CRF++}] The popular \system{CRF++}\ toolkit \cite{crf++} for conditional random fields \cite{lafferty01crf}.
\item[\system{CRF SGD}] A stochastic gradient descent conditional random field package \cite{crfsgd}.
\item[\system{Structured Perceptron}] An implementation of the structured perceptron~\cite{collins02perceptron} due to \cite{chang13svm}.
\item] The cutting-plane implementation \cite{joachims09cuttingplane} of the structured SVMs \cite{tsochantaridis04svmiso} for ``HMM'' problems.
\item[\system{Structured SVM (DEMI-DCD)}] A multicore algorithm for optimizing structured SVMs called DEcoupled Model-update and Inference with Dual Coordinate Descent.
\item[] Our approach is implemented in the Vowpal Wabbit \cite{langford07vw} toolkit on top of a cost-sensitive classifier \cite{beygelzimer05reductions} that reduces to regression trained with an online rule incorporating AdaGrad \cite{duchi2011adaptive}, per-feature normalized updates \cite{normalized}, and importance invariant updates \cite{invariant}.
\item[\system{VW Classification}] An \emph{unstructured} baseline that predicts each label independently, using one-against-all multiclass classification \cite{beygelzimer05reductions}.
\end{compactitem}

\begin{itemize}
\item  latest Vowpal Wabbit version (May 2016) commit 2dfb1225c8b89c14552932161b95237fc90b636c

\item
  CRF++ version 0.58

\item
  crfsgd version 2.0

\item
  svm\_hmm\_learn version 3.10, 14.08.08\\
    includes SVM-struct V3.10 for learning complex outputs, 14.08.08\\
    includes SVM-light V6.20 quadratic optimizer, 14.08.08

\item Illinois-SL version 0.2.2
\end{itemize}

\subsection{Hyperparameters Tuned}

The hyperparameters tuned and the values we considered for each system are:

\paragraph{CRF++}
\begin{tiny}
\begin{verbatim}
%     termination parameters:
      number of passes (--max_iter)     { 2, 4, 8, 16, 32, 64, 128 }
      termination criteria (--eta)      0.000000000001 (to prevent termination)

    tuned hyperparameters (default is *):
      learning algorithm (--algorithm)  { CRF*, MIRA }
      cost parameter (--cost)           { 0.0625, 0.125, 0.25, 0.5, 1*, 2, 4, 8, 16 }
\end{verbatim}
\end{tiny}

\paragraph{CRF SGD}
\begin{tiny}
\begin{verbatim}
%     termination parameters:
      number of passes (-r)             { 1, 2, 4, 8, 16, 32, 64, 128 }

    tuned hyperparameters (default is *):
      regularization parameter (-c)     { 0.0625, 0.125, 0.25, 0.5, 1*, 2, 4, 8, 16 }
      learning rate (-s)                { auto*, 0.1, 0.2, 0.5, 1, 2, 5 }
\end{verbatim}
\end{tiny}

\paragraph{Structured SVM}
\begin{tiny}
\begin{verbatim}
%     termination parameters:
      epsilon tolerance (-e)            { 4, 2, 1, 0.5, 0.1, 0.05, 0.01, 0.005, 0.001 }

    tuned hyperparameters (default is *):
      regularization parameter (-c)     { 0.0625, 0.125, 0.25, 0.5, 1*, 2, 4, 8, 16 }
\end{verbatim}
\end{tiny}

\paragraph{Structured Perceptron}
\begin{tiny}
\begin{verbatim}
%     termination parameters:
      number of passes (MAX_NUM_ITER)   { 1, 2, 4, 8, 16, 32, 64, 128 }

    tuned hyperparameters (default is *):
      NONE
\end{verbatim}
\end{tiny}

\paragraph{Structured SVM (DEMI-DCD)}
\begin{tiny}
\begin{verbatim}
%     termination parameters:
      number of passes (MAX_NUM_ITER)   { 1, 2, 4, 8, 16, 32, 64, 128 }

    tuned hyperparameters (default is *):
      regularization (C_FOR_STRUCTURE)  { 0.01, 0.05, 0.1*, 0.5, 1.0 }
\end{verbatim}
\end{tiny}

\paragraph{L2S}

\begin{tiny}
\begin{verbatim}
    termination parameters:
      number of passes (--passes)       { 0.01, 0.02, 0.05, 0.1, 0.2, 0.5, 1, 2, 4 }
      (note: a number of passes < 1 means that we perform one full pass, but 
       _subsample_ the training positions for each sequence at the given rate)

    tuned hyperparameters (default is *):
      base classifier                   { csoaa*}
      interpolation rate                10^{-10, -9, -8, -7, -6 }
\end{verbatim}
\end{tiny}

\paragraph{VW Classifier}
\begin{tiny}
\begin{verbatim}
    termination parameters:
      number of passes (--passes)       { 1, 2, 4 }

    tuned hyperparameters (default is *):
      learning rate (-l)                { 0.25, 0.5*, 1.0 }
\end{verbatim}
\end{tiny}

\section{Templates Used}

For part of speech tagging (CRF++):

\begin{verbatim}
    U00:%x[-2,0]
    U01:%x[-1,0]
    U02:%x[0,0]
    U03:%x[1,0]
    U04:%x[2,0]
\end{verbatim}

For named entity recognition (CRF++):

\begin{verbatim}
    U00:%x[-2,0]
    U01:%x[-1,0]
    U02:%x[0,0]
    U03:%x[1,0]
    U04:%x[2,0]
    
    U10:%x[-2,1]
    U11:%x[-1,1]
    U12:%x[0,1]
    U13:%x[1,1]
    U14:%x[2,1]
    
    U15:%x[-2,1]/%x[-1,1]
    U16:%x[-1,1]/%x[0,1]
    U17:%x[0,1]/%x[1,1]
    U18:%x[1,1]/%x[2,1] 
\end{verbatim}

    (where words are in position 0 and POS is in 1)

Additional features for L2S (ft) on POS Tagging:

\begin{verbatim}
  -- the left and the right tokens of each word
  -- the first and the last 2 characters for each token
\end{verbatim}

For L2S (ft) on NER:

\begin{verbatim}
  -- the left and the right two tokens of each word
  -- the POS tags of the left and the right tokens for each word
  -- the last charaster for each token
\end{verbatim}

 \end{document}

%% file: preamble.tex
\usepackage{hyperref}
\usepackage{url}
\usepackage[T1]{fontenc}
\usepackage[latin9]{inputenc}
\usepackage{float}
\usepackage{amsthm}
\usepackage{multirow}
\usepackage{xcolor}
\usepackage{graphicx}
\usepackage{algorithm}
\usepackage{algorithmic}
\usepackage{amsthm}

\definecolor{mydarkblue}{rgb}{0,0.08,0.45}
\hypersetup{ %
  pdftitle={},
  pdfauthor={Anonymous Submission},
  pdfsubject={},
  pdfkeywords={},
  pdfborder=0 0 0,
  pdfpagemode=UseNone,
  colorlinks=true,
  linkcolor=mydarkblue,
  citecolor=mydarkblue,
  filecolor=mydarkblue,
  urlcolor=mydarkblue,
  pdfview=FitH}

\newtheorem{theorem}{Theorem}
\newtheorem{definition}{Definition}

\algsetup{linenosize=\tiny,
          linenodelimiter=:
          }

\renewcommand{\cite}[1]{\citep{#1}}

\definecolor{darkgrey}{rgb}{0.2,0.2,0.2}
\definecolor{grey}{rgb}{0.9,0.9,0.9}
\definecolor{darkblue}{rgb}{0.0,0.0,0.5}
\definecolor{darkpurple}{rgb}{0.4,0.0,0.4}
\definecolor{darkred}{rgb}{0.5,0.0,0.0}
\definecolor{darkorange}{rgb}{0.6,0.4,0.1}
\definecolor{darkgreen}{rgb}{0.0,0.5,0.0}
\definecolor{darkergreen}{rgb}{0.0,0.4,0.0}
\definecolor{lightblue}{rgb}{0.8,0.8,1.0}
\definecolor{lightgreen}{rgb}{0.8,1.0,0.8}
\definecolor{lightred}{rgb}{1.0,0.8,0.8}
\definecolor{lightyellow}{rgb}{1.0,1.0,0.8}
\definecolor{lightorange}{rgb}{1.0,0.9,0.8}
\definecolor{lightgrey}{rgb}{0.96,0.97,0.98}

\newcommand{\wordtag}[1]{\textsf{\scriptsize{\textcolor{darkblue}{#1}}}}
\newcommand{\overbracetxt}[2]{$\textcolor{darkblue}{\overbrace{\textnormal{\textcolor{black}{#1}}}^{\textsf{#2}}}$}

\newcommand{\algorithmicfont}{\normalfont\sffamily\footnotesize}
\newcommand{\algorithmiccolor}{darkergreen}

\newcommand{\VAR}[1]{\textcolor{darkblue}{\textit{#1}}}
\newcommand{\CON}[1]{\textcolor{darkgrey}{\textit{#1}}}
\newcommand{\FUN}[1]{\textcolor{darkpurple}{\textsc{#1}}}

\newcommand{\SETST}[1]{\STATE \VAR{#1} $\leftarrow$ }

\floatname{algorithm}{\textcolor{\algorithmiccolor}{\algorithmicfont Algorithm}}

\newcommand{\newalgorithm}[3]{%
  \begin{algorithm}[t]
    \caption{#2}
    \begin{small}
      \begin{algorithmic}[1]
        #3
      \end{algorithmic}
    \end{small}
    \label{alg:#1}
  \end{algorithm}}

\newcommand{\predict}{\FUN{predict}(...)}
\newcommand{\loss}{\FUN{loss}(...)}

\usepackage{babel}

\renewcommand{\brack}[1]{\left\{\begin{array}{ll}#1\end{array}\right.}
\newcommand{\myreturn}{\textcolor{\algorithmiccolor}{\algorithmicfont return~~}}
\newcommand{\cO}{\mathcal{O}}

\newcommand{\system}[1]{\textsf{\footnotesize #1}}

\newcommand{\textbsf}[1]{{\bf \textsf{\small #1}}}